\pdfoutput=1

\documentclass[11pt]{article}

\usepackage{acl}

\def\code#1{\texttt{#1}}
\usepackage{xspace}

\usepackage{array, boldline, makecell, booktabs}
\usepackage[svgnames, table]{xcolor}
\newcommand{\thickhline}{\hlineB{4}}

\makeatletter
\NewDocumentCommand{\supptitle}{s}{
\twocolumn[{
  \begin{center}
    \vspace*{-0.3cm}
    \rule{\textwidth}{0.05cm}\\[0.1cm]
    \textbf{- Appendix -}\\[0.2cm]
    {\Large \textbf{\mytitle}}\\[0.1cm]
    \rule{\textwidth}{0.05cm}\\[0.3cm]
  \end{center}
}]
}
\makeatother

\newcommand{\eg}{\emph{e.g.,~}}
\newcommand{\ie}{\emph{i.e.,~}}

\definecolor{LightCyan}{rgb}{0.88,1,1}
\definecolor{Blue}{rgb}{0, 0.5, 1}
\definecolor{Green}{rgb}{0.0, 0.8, 0.0 }
\definecolor{Red}{rgb}{0.95, 0.55, 0.6}
\definecolor{Skyblue}{rgb}{0.6, 0.6, 0.95 }
\definecolor{Beige}{rgb}{0.96, 0.96, 0.86}

\newcommand{\alg}{\code{SCoNE}\xspace}
\newcommand{\mytitle}{SCoNE: Selective Context-aware Neuron Editing for Robust Retrieval-Augmented Generation}

\usepackage[normalem]{ulem}
\useunder{\uline}{\ul}{}

\usepackage{tcolorbox}
\tcbuselibrary{skins, breakable}

\usepackage[hang,flushmargin]{footmisc}

\usepackage{amsmath,amsfonts,bm}

\def\eqref#1{equation~\ref{#1}}

\def\1{\bm{1}}

\DeclareMathAlphabet{\mathsfit}{\encodingdefault}{\sfdefault}{m}{sl}
\SetMathAlphabet{\mathsfit}{bold}{\encodingdefault}{\sfdefault}{bx}{n}

\usepackage{times}
\usepackage{latexsym}
\usepackage{tikz}
\usetikzlibrary{positioning, arrows.meta, shapes.geometric, calc, decorations.pathreplacing}

\usepackage{amsmath}  
\usepackage{amssymb}  
\usepackage{xcolor}  
\usepackage[T1]{fontenc}

\usepackage[utf8]{inputenc}

\usepackage{microtype}

\usepackage{inconsolata}

\usepackage{graphicx}

\usepackage{algorithm}
\usepackage{footmisc}

\usepackage{newfloat}
\usepackage{listings}
\DeclareCaptionStyle{ruled}{labelfont=normalfont,labelsep=colon,strut=off} 
\floatstyle{ruled}
\newfloat{listing}{tb}{lst}{}
\floatname{listing}{Listing}
\usepackage{times}
\usepackage{latexsym}
\usepackage{mathtools}
\usepackage{amsthm}

\usepackage[capitalize,noabbrev]{cleveref}

\usepackage{tikz}
\usetikzlibrary{fadings}
\usetikzlibrary{patterns}
\usetikzlibrary{shadows.blur}
\usetikzlibrary{shapes}

\usepackage{latexsym}
\usepackage{nccmath}
\usepackage{bbm}
\usepackage{tabularx}
\usepackage{yhmath}
\usepackage{times}
\usepackage{latexsym}
\usepackage{booktabs}
\usepackage{graphicx}
\usepackage{algorithm}
\usepackage{subcaption}
\usepackage{algpseudocode}
\usepackage{amsmath}
\usepackage{siunitx}
\usepackage{mathtools}
\usepackage{array}
\usepackage{multirow}
\usepackage[normalem]{ulem}
\usepackage{arydshln}

\useunder{\uline}{\ul}{}
 
\makeatletter
\newcommand{\multiline}[1]{%
  \begin{tabularx}{\dimexpr\linewidth-\ALG@thistlm}[t]{@{}X@{}}
    #1
  \end{tabularx}
}
\makeatother

\algblock{Input}{EndInput}
\algnotext{EndInput}
\algblock{Output}{EndOutput}
\algnotext{EndOutput}

\title{\mytitle}

\author{
  \textbf{Chaewon Kim} \quad \textbf{Seo Yeon Park}\thanks{Corresponding author} \\
  Hanyang University \\
  \texttt{\{rud14dns, seoyeonpark\}@hanyang.ac.kr}
}

\begin{document}

\maketitle

\begin{abstract}
Retrieval-Augmented Generation (RAG) is highly sensitive to retrieval noise: when retrieved documents mix informative and irrelevant context, LLMs are easily distracted, leading to hallucinations. To overcome this, we propose \alg (Selective Context-aware Neuron Editing), a training-free model editing approach that improves retrieval noise robustness by selectively strengthening context-aware FFN neurons that are identified by both high attribution and high cross-input variability. \alg requires only a small number of mining samples, no fine-tuning, and no inference-time overhead. Across various knowledge-intensive question-answering benchmarks and two LLM backbones, \alg consistently outperforms competitive baseline methods. 
Our code is available at \url{https://github.com/HYU-ARK-Lab/SCoNE}.

\end{abstract}

\section{Introduction}
\label{sec:intro}

While Large Language Models (LLMs) have achieved remarkable success, they remain prone to hallucinations in knowledge-intensive tasks~\cite{wang-yu-2025-iquest, huang2025survey}. Retrieval-Augmented Generation (RAG)~\citep{lewis2020retrieval} mitigates this by grounding outputs in externally retrieved evidence. However, the effectiveness of RAG is heavily dependent on the quality of retrieved documents, which retrieval systems cannot always guarantee. In realistic settings, retrievers return a mixture of relevant, partially relevant, and irrelevant documents for the same query, and LLMs are known to be easily distracted by such \emph{retrieval noise}, often degrading rather than improving their answers~\citep{yoran2024makingretrievalaugmentedlanguagemodels, shi2023large}.

Existing approaches to retrieval noise robustness span several paradigms, including prompt engineering, retrieved-context refinement and {fine-tuning}. 
While solutions such as introducing additional modules (\eg reranker, compressor) are flexible, they introduce additional components into the RAG pipeline, which can lead to cascading errors across each stage~\citep{asai2024self, yoran2024makingretrievalaugmentedlanguagemodels} and substantial inference-time latency~\citep{an2025hyperrag}. 
In contrast, directly fine-tuning the generator to be robust against retrieval noise avoids such pipeline overhead and has therefore emerged as a promising direction \cite{yoran2024makingretrievalaugmentedlanguagemodels, wu-etal-2025-pa}.  
However, fine-tuning-based methods inherit the well-known drawbacks of gradient-based adaptation: catastrophic forgetting, substantial compute requirements, and the need for carefully curated training data. A natural question arises: \emph{can retrieval noise robustness be achieved without retraining the model?}

Model editing offers a promising alternative paradigm. By directly modifying a small number of parameters, editing methods provide fine-grained control over model behavior without the cost of fine-tuning. 
However, existing model editing methods fundamentally assume that the target knowledge to be edited is known in advance \cite{NEURIPS2022_6f1d43d5, meng2022mass}. This assumption does not hold in Retrieval-Augmented Generation (RAG), where retrieved contexts are inherently open-ended and dynamically vary across queries. In RAG, the model cannot anticipate which facts will appear in the retrieved documents at inference time.
Hence, rather than fixing a specific parametric knowledge, RAG requires an adaptive approach to constrain the model's behavior in response to whatever context is retrieved at inference, regardless of its specific content. This challenge is further complicated by the realistic nature of retrieval itself; even for a single query, some documents directly support the answer, others are partially relevant, and others are entirely irrelevant. For this, a method that not only makes the model responsive to context, but selectively responsive to a context (\ie engaging with informative documents while remaining unaffected by noisy ones) is necessarily required. 

Building on this insight, we propose \textbf{\alg} (\textbf{S}elective \textbf{Co}ntext-aware \textbf{N}euron \textbf{E}diting) for RAG, which is a model editing method to enhance robustness for retrieval noise. \alg identifies \emph{selectively context-aware neurons} by jointly requiring high attribution and high cross-input variability, and strengthens them at inference time.
Our method requires only 100 mining samples from a single dataset (HotpotQA), no fine-tuning, and no inference-time overhead beyond standard RAG. Across various benchmarks, \alg consistently outperforms strong competitive baselines, demonstrating that lightweight editing, when guided by the right neuron selection criterion, can match or exceed the effectiveness of heavyweight fine-tuning pipelines. 
Similarly, \citet{shi2024ircan} adopt a knowledge-agnostic approach but identify context-aware neurons based on attribution strength alone. 
While this criterion is effective in their single-context setting, RAG presents multiple retrieved documents containing both informative and distracting evidence simultaneously. Here, a neuron may receive high attribution simply because it responds broadly to any retrieved content, potentially reflecting sensitivity to surface-level patterns rather than informative evidence.
Thus, attribution strength alone is insufficient for identifying neurons that specifically mediate useful evidence utilization in noisy retrieval settings.
Our redefinition of context-aware neurons addresses this gap by complementing attribution strength with cross-input variability, making the criterion more suitable for RAG's heterogeneous, multi-document setting.

\section{Method}

\paragraph{Problem Setup.} 
Consider a neuron mining dataset of size $N$, sampled from the training split of HotpotQA~\citep{yang2018hotpotqa}, $\mathcal{D}=\{(q,\mathcal{C},a)_t\}_{t=1}^{N}$, where the $t$-th instance consists of a query $q_t$, its ground-truth answer $a_t$, and an associated context set $\mathcal{C}_t = \{c^g_m\}_{m=1}^{M} \cup \{c^d_k\}_{k=1}^{K}$. Here, $\{c^g_m\}_{m=1}^{M}$ denote the gold contexts that directly support $a_t$, while $\{c^d_k\}_{k=1}^{K}$ are distractor contexts that do not entail the answer. 
Our goal is to characterize, for each instance, how individual FFN neurons engage with such mixed evidence during inference. To this end, we define two complementary measures, \emph{attribution} and \emph{variability}, computed over the course of inference.

\subsection{Neuron Mining}
\label{sec:gen}

\paragraph{Attribution.}
Prior work has shown that factual knowledge is localized in specific FFN neurons~\citep{dai-etal-2022-knowledge,  geva-etal-2021-transformer}. We hypothesize that neurons responsible for processing retrieved contextual information also reside in FFNs, and seek to identify those that engage with retrieved evidence during RAG inference. Following \citet{shi2024ircan}, we estimate the attribution of each FFN neuron $n$ via Integrated Gradients~\cite{sundararajan2017axiomatic}, but extend their single-context formulation to the multi-context RAG setting where gold and distractor evidence co-occur. 
For each instance $(q, \mathcal{C}, a)_t \in \mathcal{D}$, we quantify how individual FFN neurons contribute to the model's prediction when the query is presented together with its full context set $\mathcal{C}_t = \{c^g_m\}_{m=1}^{M} \cup \{c^d_k\}_{k=1}^{K}$, which contains both gold and distractor contexts. Specifically, we formulate the attribution score calculation as follows: let $v(q_t)$ denote the activation of neuron $n$ when the model is given the query alone, and let $v(q_t, \mathcal{C}_t)$ denote its activation when the query is augmented with the full context set $\mathcal{C}_t$. The attribution score is then defined as follows:

\begin{equation}
\begin{split}
    \text{Attr}(n;\, q_t, \mathcal{C}_t)
    &= \bigl(v(q_t, \mathcal{C}_t) - v(q_t)\bigr) \\
    &\quad \times \int_{\alpha=0}^{1}
    \frac{\partial\, P(a \mid q_t, \mathcal{C}_t, v_{\alpha})}
         {\partial\, v_{\alpha}}\, d\alpha,
\end{split}
\label{eq:attr}
\end{equation}
where $v_{\alpha} = v(q_t) + \alpha\bigl(v(q_t, \mathcal{C}_t) - v(q_t)\bigr)$ linearly interpolates between the query-only and the query-plus-context activations for $\alpha \in [0, 1]$. In practice, the integral is approximated by a 20-step Riemann sum.

\begin{table*}[ht]
\centering
\resizebox{0.95\textwidth}{!}{
\begin{tabular}{lcccccccc}
\toprule
& \multicolumn{1}{c}{\textbf{NQ}} & \multicolumn{1}{c}{\textbf{ASQA}} & \multicolumn{1}{c}{\textbf{SCIQ}} &  \multicolumn{1}{c}{\textbf{TriviaQA}} & \multicolumn{1}{c}{\textbf{HQA}} & \multicolumn{1}{c}{\textbf{TruthfulQA}} & \multicolumn{1}{c}{\textbf{PopQA}} & \multicolumn{1}{c}{\multirow{1}{*}{\shortstack{\textbf{Avg.}}}} \\ \hline
\multicolumn{9}{l}{\textbf{\textit{Llama-3-8B-Instruct}}} \\ \hline
\thickhline 

\code{RAG} \cite{lewis2020retrieval} & 62.81  & 68.78 & 54.10 & 88.65 & 46.55  & 4.90  & 60.17  & 55.14 \\  \hline\hline

\code{RetRobust}~\cite{yoran2024makingretrievalaugmentedlanguagemodels}
& 62.71  & 69.62 & 53.10   & 88.53  & 46.39 & 5.14  & 60.64 &  55.16 \\ \hline
\code{PA-RAG} \cite{wu-etal-2025-pa}
& \textbf{68.06}  & \underline{73.73} & \underline{56.80}  & \underline{90.18}  & \underline{50.41} & 3.79 & \underline{64.19}  & \underline{58.17} \\ \hline \hline
\code{CAD} \cite{shi-etal-2024-trusting}
& 64.12 & 68.99 & 48.50 & 87.81  & 46.77 & 3.55   & 62.21 & 54.56 \\ \hline
\code{IRCAN} \cite{shi2024ircan}
& 64.65 & 71.41 & 54.00  & 89.67 & 50.11 & \underline{5.51}  & 63.65 & 57.00 \\ \hline\hline
\rowcolor{LightCyan}
\textbf{\alg} (Ours)
& \underline{66.44} & \textbf{73.84} & \textbf{57.10}  & \textbf{90.66} & \textbf{52.27} & \textbf{6.36}  & \textbf{65.57}     & \textbf{58.89}  \\ \hline

\toprule
\multicolumn{9}{l}{\textbf{\textit{Qwen-2.5-7B-Instruct}}} \\ \hline
\thickhline

\code{RAG} \cite{lewis2020retrieval}     
& \underline{62.25} & 69.09 & \underline{54.90} & 86.98 & 45.54 & \textbf{6.24} & 58.54 & \underline{54.79} \\ \hline\hline

\code{RetRobust} \cite{yoran2024makingretrievalaugmentedlanguagemodels}
& 60.73 & 67.19  & 53.90 & 86.83  & 44.41 & 5.51  & 57.48 &  53.72 \\ \hline
\code{PA-RAG} \cite{wu-etal-2025-pa}
& 60.45  & 68.35 & 52.90  & \underline{87.54}  & \textbf{48.00} & \underline{6.00} & 56.65  & 54.27 \\ \hline \hline
\code{CAD} \cite{shi-etal-2024-trusting}     
& 61.83 & \underline{69.62} & 51.50 & 85.62 & 43.88  & 4.04 & \underline{58.84} & 53.62 \\ \hline
\code{IRCAN}~\cite{shi2024ircan}
& \underline{62.25} & 68.78 & 54.20 & 87.28 &{45.86} & \underline{6.00} & 58.82 & 54.74 \\ \hline 
\rowcolor{LightCyan} 
\textbf{\alg} (Ours)
& \textbf{63.24} & \textbf{69.94} & \textbf{56.20} & \textbf{88.16} & \underline{47.00} & {5.63} & \textbf{60.04} & \textbf{55.74} \\ \hline 
\thickhline
\end{tabular}
}
\vspace{-8pt}
\caption{Accuracy comparison across QA benchmarks using Llama-3-8B-Instruct (top) and Qwen-2.5-7B-Instruct (bottom). \textbf{Bold} scores are best in each dataset, and \underline{underlined} scores are the second-best results.}
\vspace{-8pt}
\label{tb:main}
\end{table*}

\paragraph{Variability.} 
Attribution alone, however, cannot distinguish between two qualitatively different neuron behaviors: neurons that selectively respond to specific contexts and neurons that activate uniformly regardless of context content. Only the former captures the content-level selectivity required in the heterogeneous multi-document setting of RAG. 
To capture this context selectivity, we measure how a neuron’s attribution varies across different query-context instances. 
Concretely, let $\mathrm{Attr}^{(t)}(n^l_j)$ denote the attribution of the $j$-th intermediate neuron in the $l$-th FFN layer for the $t$-th instance $(q,\mathcal{C},a)_t \in \mathcal{D}$. The variability score is defined as the deviation of the current attribution from its running average over the preceding $W$ instances:

{\small
\begin{equation}
V^{(t)}(n^l_j) = \left| \mathrm{Attr}^{(t)}(n^l_j) - \frac{1}{W}\sum_{m=t-W}^{t-1} \mathrm{Attr}^{(m)}(n^l_j) \right|
\label{eq:var}
\end{equation}
}
\noindent where $W$ denotes the number of preceding instances used to compute the running average. We compute this score over a fixed traversal order of the mining set $\mathcal{D}$, so that the sliding window captures local variation in the neuron's attribution across neighboring instances in the traversal.

\paragraph{High-Attribution and Variability Neuron Selection.} 
Neurons with both high attribution and high variability are considered context-aware: they contribute strongly to the current context while responding differently across inputs. 
Hence, we select the neurons as follows: For each sample $(q,\mathcal{C},a)_t$, we construct $\mathcal{A}^{(t)}$ and $\mathcal{B}^{(t)}$, containing the top-50 neurons ranked by attribution and variability, respectively, restricted to neurons with positive attribution scores.\footnote{Restricting to positive attribution prevents neurons whose variability stems from transitions between negative and near-zero attribution, whose overall contribution remains negligible, from being selected.} Their intersection $\mathcal{C}^{(t)} = \mathcal{A}^{(t)} \cap \mathcal{B}^{(t)}$ forms the locally selected neurons for sample $t$. We then aggregate the selection frequency of each neuron across all samples and choose the top-$k$ most frequent neurons as the final context-aware neuron set.
For scale, we treat each layer-specific FFN dimension as a distinct neuron. 
Llama-3-8B-Instruct contains $32 \times 14{,}336 = 458{,}752$ such neurons. Each of the top-50 sets $\mathcal{A}^{(t)}$ and $\mathcal{B}^{(t)}$ corresponds to $\approx 0.0109\%$ of all layer-specific FFN neurons. 
With $k=5$, the final neuron set therefore contains $\approx 0.0011\%$ of all layer-specific FFN neurons.

\subsection{Neuron Enhancement}
Once context-aware neurons are identified, we amplify their contribution at inference time to better leverage informative retrieved evidence. For each selected neuron $n_i^l$, we scale its corresponding FFN weight as follows: $\hat{W}(n_i^l) = \alpha \cdot W(n_i^l)$, where $\alpha$ controls the enhancement strength.

\section{Experimental Setup}
\vspace{-2mm}

\paragraph{Neuron Mining Dataset.} 
{We sample the first 100 instances}
from the HotpotQA~\cite{yang2018hotpotqa} training split for neuron mining. We set the number of gold-content $M=2$, and the distractor context $K=8$. 

\paragraph{Evaluation Dataset.} We use the dev splits provided by BERGEN~\cite{rau-etal-2024-bergen} for NQ~\cite{kwiatkowski2019natural}, ASQA~\cite{stelmakh2022asqa}, SCIQ~\cite{welbl2017crowdsourcing}, TriviaQA~\cite{joshi2017triviaqa}, HotpotQA~\cite{yang2018hotpotqa}, TruthfulQA~\cite{lin-etal-2022-truthfulqa}, PopQA~\cite{mallen-etal-2023-trust}. All retrieval documents are sourced from the KILT~\cite{petroni-etal-2021-kilt} Wikipedia dump\footnote{\url{https://huggingface.co/datasets/facebook/kilt\_wikipedia}}, and we retrieve top-5 documents per question using SPLADE-v3~\cite{lassance2024splade}.

\paragraph{Baselines.}
\vspace{-2mm}
We compare \alg against representative RAG baselines: \code{RAG}~\cite{lewis2020retrieval}; 
\code{RetRobust}~\cite{yoran2024makingretrievalaugmentedlanguagemodels} and \code{PA-RAG}~\cite{wu-etal-2025-pa} for generator fine-tuning; and \code{CAD}~\cite{shi-etal-2024-trusting} and \code{IRCAN}~\cite{shi2024ircan} for inference-time intervention at the decoding and parameter level, respectively.

\vspace{-2mm}
\paragraph{Implementation Details.}

Our experiments are performed using the RAG framework provided by BERGEN~\cite{rau-etal-2024-bergen}, which offers a realistic RAG pipeline. We use Llama-3-8B-Instruct~\cite{dubey2024llama3herdmodels} and Qwen-2.5-7B-Instruct \cite{yang2024qwen2} as the generator LLM, and SPLADE-v3~\cite{lassance2024splade} as the retriever.
For retrieval, we use the KILT Wikipedia dump\footnote{https://huggingface.co/datasets/kilt\_wikipedia}, preprocessed into non-overlapping 100-word chunks, and retrieve five documents per question. 
All experiments are conducted on a single NVIDIA H200 GPU. 
For fair comparison, both IRCAN and our method identify neurons from the same neuron mining dataset $\mathcal{D}$. 
Details of $\mathcal{D}$ are provided in \ref{app:neuron selection}. We set the enhancement strength $\alpha = 7$, select the top-$k$ context-aware neurons where $k = 5$, and the window size $W=3$.

\begin{table}[t]
\centering
\resizebox{1.00\columnwidth}{!}{
\begin{tabular}{lcccccc}   
\thickhline
& \multicolumn{3}{c}{\textbf{Relevant}} & \multicolumn{3}{c}{\textbf{Irrelevant}} \\
\cmidrule(lr){2-4} \cmidrule(lr){5-7}
\textbf{Llama-3-8B-Instruct} & NQ & SCIQ & HQA & NQ & SCIQ & HQA \\ \hline
\code{RAG} \cite{lewis2020retrieval}     
& 78.04 & 73.61 & 72.75 & 4.43 & \textbf{8.36} & 15.16 \\ \hline
\code{PA-RAG} \cite{wu-etal-2025-pa} 
& \textbf{85.29} & \textbf{78.60} & \textbf{81.26} & 2.04 & 5.69 & 13.43 \\ \hline
\code{IRCAN}~\cite{shi2024ircan}
& 80.44 & 73.75 & 76.88 & 4.09 & 7.69 & 18.02 \\ \hline 
\rowcolor{LightCyan} 
\textbf{\alg} (Ours)
& 82.00 & 78.03 & 79.53 & \textbf{6.81} & 8.03 & \textbf{19.59} \\ \hline 
\thickhline
\end{tabular}
}
\vspace{-8pt}
\caption{The comparison of accuracy on Relevant and Irrelevant subsets.}
\vspace{-8pt}
\label{tb:irrelevant}
\end{table}

\section{Results}
\vspace{-2mm}
\paragraph{Main Results.} 
Table~\ref{tb:main} reports the main results. Accuracy is measured using the Match score, which checks whether the gold answer appears as a substring of the generated output. 
\alg achieves the best overall performance with Llama-3-8B-Instruct, ranking first on six of seven datasets and improving over \code{RAG} by 3.75\% on average. 
Compared to fine-tuning baselines, \alg outperforms \code{RetRobust} by {3.73}\% and remains within 0.72\% of \code{PA-RAG} despite requiring no additional training. 
Against intervention-based baselines, \alg surpasses \code{CAD} on all datasets and improves over \code{IRCAN} by up to 3.1\% on SCIQ using Llama-3-8b-Instruct,  suggesting that our variability-based criterion identifies neurons more selectively responsive to retrieved context than attribution alone. 
A similar trend holds with Qwen-2.5-7B-Instruct as the generator, where \alg consistently outperforms \code{IRCAN} across most benchmarks. 
\alg achieves the best average accuracy, demonstrating its effectiveness across different generators. 
{
This advantage is also preserved under LLM-based evaluation (Appendix ~\ref{app:llm_eval}).
}
We confirm that \alg's improvements stem from neuron selection: randomly selected neurons remain on par with vanilla RAG (Appendix~\ref{app:random_neuron}).
We further verify that these gains are robust to the choice of mining sample, with accuracy remaining stable across different 100-example samples from HotpotQA (Appendix~\ref{app:mining_samples}).

\begin{table}[t]
\centering
\small
\resizebox{\linewidth}{!}{%
\begin{tabular}{lccc}
\toprule
\textbf{Measure} & \textbf{NQ} & \textbf{SCIQ} & \textbf{HQA} \\
\midrule
Variance / Std. & 63.52 & 54.70 & 50.25 \\
Mean Absolute Deviation (MAD) & 64.61 & 54.10 & 50.27 \\
\alg & 66.44 & 57.10 & 52.27 \\
\bottomrule
\end{tabular}%
}
\caption{The comparison of the proposed variability with order-invariant measures for neuron selection on Llama-3-8B-Instruct.}
\label{tab:variability_measure}
\end{table}

\paragraph{Relevant vs. Irrelevant Context Analysis.}
To examine whether our method exhibits context-dependent behavior with retrieved contexts, we divide each evaluation set into two subsets:
\textit{Relevant}, where at least one retrieved document contains the gold answer, and \textit{Irrelevant}, otherwise.
As shown in Table~\ref{tb:irrelevant}, \alg consistently improves over \code{RAG} and \code{IRCAN} on the relevant subset across all datasets. 
While \code{PA-RAG} achieves the highest accuracy on the relevant subset, \alg demonstrates stronger robustness under irrelevant contexts, outperforming all baselines, including vanilla \code{RAG}, on NQ and HQA.
This robustness holds under a controlled noise experiment (Appendix~\ref{app:controlled_noise}). 
Notably, \alg surpasses \code{IRCAN} on both the Relevant and Irrelevant subsets, suggesting that incorporating cross-input variability beyond attribution strength helps identify neurons that selectively engage with informative evidence. 
This selectivity is reflected in their activation patterns across different retrieved-evidence compositions (Appendix~\ref{app:neuron_validation}).

\paragraph{Comparison of Variability Measures}
Our variability measure in Eq.~\ref{eq:var} uses an unsigned running residual and thus depends on example order.
We compare it with three order-invariant alternatives---variance, standard deviation, and mean absolute deviation (MAD)---computed over the full set of attribution scores for each neuron, irrespective of their traversal order.
With all other settings fixed, Table~\ref{tab:variability_measure} shows that \alg consistently outperforms these measures across all three datasets on Llama-3-8B-Instruct.
Variance and standard deviation yield identical results because they induce the same neuron ranking and select the same top-5 neurons.
Overall, the results suggest that \alg's running-residual formulation provides a more effective variability signal for identifying selective context-aware neurons.

\paragraph{Ablation on Neuron Selection Criteria}
To isolate the contribution of cross-input variability beyond attribution alone, we compare three neuron-selection strategies while keeping all other settings identical: Attr-only, Var-only, and Attr+Var (SCoNE).
In Table~\ref{tab:selection_ablation} Attr+Var consistently outperforms Attr-only by 1.83, 3.00, and 2.00\% on NQ, SCIQ, and HotpotQA, respectively. 
Var-only is insufficient by itself, whereas its combination with attribution consistently yields the best performance. 
This demonstrates that variability provides a complementary and substantive signal for neuron identification.

\begin{table}[t]
\centering
\small
\begin{tabular}{lccc}
\toprule
\textbf{Selection} & \textbf{NQ} & \textbf{SCIQ} & \textbf{HotpotQA} \\
\midrule
Attr-only & 64.61 & 54.10 & 50.27 \\
Var-only & 63.52 & 54.70 & 50.25 \\
Attr+Var (\alg) & \textbf{66.44} & \textbf{57.10} & \textbf{52.27} \\
\bottomrule
\end{tabular}
\vspace{-2mm}
\caption{The comparison of neuron selection criteria on Llama-3-8B-Instruct.}
\label{tab:selection_ablation}
\end{table}

\begin{figure}[t]
    \centering
    \includegraphics[width=\columnwidth]{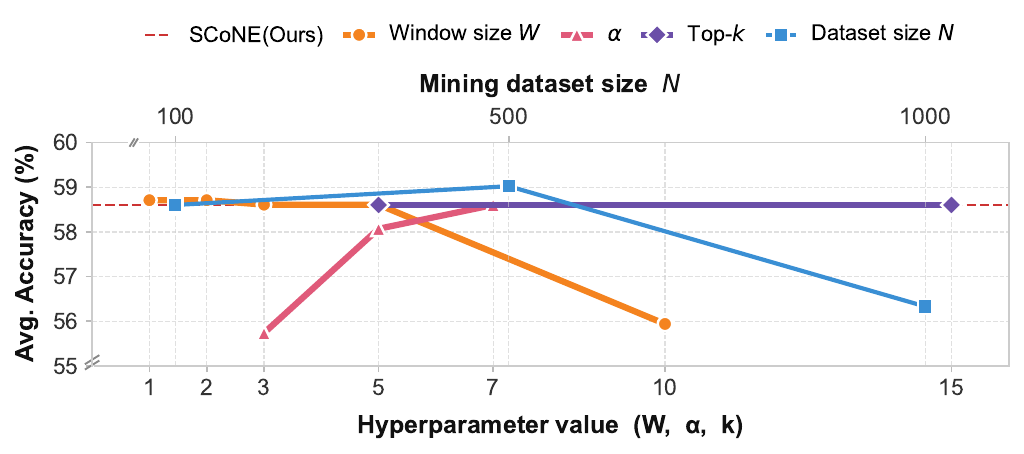}
    \vspace{-8mm}
    \caption{
    Ablation study on key hyperparameters using Llama-3-8B-Instruct. Results are averaged across NQ, SCIQ, and HotpotQA.
    }
    \label{fig:ablation}
    \vspace{-2mm}
\end{figure}

\paragraph{Hyperparameter Analysis.}
We conduct ablation studies on the enhancement strength $\alpha$, context window size $W$, neuron mining dataset size $N$, and the number of selected neurons $k$ using Llama-3-8B-Instruct. The result is shown in Figure~\ref{fig:ablation}.\footnote{More detailed results for each hyperparameter setting are provided in \ref{effect_neuron}, \ref{effect window}, \ref{effect dataset}.}
Performance improves monotonically with $\alpha$, peaking at $\alpha=7$, and remains stable across small-to-moderate $W,N$, and $k$.
Overall, while extreme hyperparameter values (\emph{e.g.}, $W=10$ or $N=1000$) degrade performance, the default \alg configuration denoted as \textit{SCoNE (Ours)} consistently achieves the best or near-best performance, validating our design choices.\footnote{\alg(Ours) corresponds to the default configuration with $\alpha=7$, $W=3$, $N=100$, and $k=5$.}

\section{Related Work}
\label{sec:related_work}

\paragraph{Retrieval Noise Robustness in RAG.}
Prior work has addressed retrieval noise in Retrieval-Augmented Generation (RAG) through prompt engineering~\citep{zhou-etal-2023-context}, analyses of retrieved-context composition~\citep{powerofnoise}, retrieved-context refinement, and fine-tuning.
Retrieved-context refinement includes reranking and compression~\citep{glass-etal-2022-re2g, xu2023recomp}, with recent approaches further exploring compact clue selection~\citep{zhang2026less}, reinforcement-learning-based evidence extraction~\citep{zhao-etal-2026-learning-extract}, and attention-based context compression~\citep{zhang2026sentineldecodingcontextutilization}.
Fine-tuning approaches instead adapt the generator itself to improve robustness against retrieval noise.
\citet{yoran2024makingretrievalaugmentedlanguagemodels} train the generator on mixtures of relevant and irrelevant contexts, while \citet{wu-etal-2025-pa} align it via multi-perspective preference optimization.
More recently, \citet{conflictrag} incorporate conflict signals into multi-stage learning to improve robustness against conflicting retrieved knowledge.

\paragraph{Model Editing for Context Utilization.}
Model editing modifies model parameters to alter knowledge or behavior without additional training.
Methods typically target specific knowledge known in advance~\citep{NEURIPS2022_6f1d43d5,meng2022mass}.
Recent studies have extended model-level interventions toward knowledge-agnostic control of contextual knowledge utilization.
\citet{shi2024ircan} perform neuron-level model editing by identifying and reweighting context-aware neurons based on attribution strength, enabling knowledge-agnostic adaptation to contextual knowledge.
\alg builds on this knowledge-agnostic, neuron-level perspective for noisy multi-document RAG, where informative and distracting contexts coexist.
It therefore complements attribution strength with cross-input variability to identify selectively context-responsive neurons.

\vspace{-1mm}
\section{Conclusion}

We present \alg, a framework that improves RAG robustness to retrieval noise by selectively enhancing context-aware neurons, identified through attribution strength and cross-input variability. Experiments across various benchmarks show that \alg matches or surpasses strong baselines, demonstrating that variability-based neuron mining provides a practical criterion. 
\vspace{-1mm}

\section*{Limitations}
While the selected neurons transfer effectively across diverse benchmarks, several limitations remain. 
In this work, neuron mining is performed using samples from HotpotQA only, and it remains unclear how the characteristics of the mining dataset influence the selected neuron set and downstream behavior. For example, using more challenging QA datasets may lead to different neuron distributions and transfer properties.
In addition, our current setting relies on contexts containing both gold-content and distractor-content. Therefore, it remains unclear how neuron selection would differ under cleaner retrieval settings containing only gold supporting documents. Investigating how mining dataset composition and retrieval conditions affect neuron mining and robustness remains an important direction for future work.

\section*{Acknowledgments}
This work was supported by the National Research Foundation of Korea (NRF) grant funded by the Korea government (MSIT) (RS-2025-24535182 and RS-2026-25498006).

\bibliography{custom}

\begin{thebibliography}{39}
\providecommand{\natexlab}[1]{#1}

\bibitem[{An et~al.(2025)An, Cheng, Park, and Jiang}]{an2025hyperrag}
Yuwei An, Yihua Cheng, Seo~Jin Park, and Junchen Jiang. 2025.
\newblock Hyperrag: Enhancing quality-efficiency tradeoffs in retrieval-augmented generation with reranker kv-cache reuse.
\newblock \emph{arXiv preprint arXiv:2504.02921}.

\bibitem[{Asai et~al.(2024)Asai, Wu, Wang, Sil, and Hajishirzi}]{asai2024self}
Akari Asai, Zeqiu Wu, Yizhong Wang, Avirup Sil, and Hannaneh Hajishirzi. 2024.
\newblock \href {https://openreview.net/forum?id=hSyW5go0v8} {Self-{RAG}: Learning to retrieve, generate, and critique through self-reflection}.
\newblock In \emph{The Twelfth International Conference on Learning Representations}.

\bibitem[{Clark et~al.(2018)Clark, Cowhey, Etzioni, Khot, Sabharwal, Schoenick, and Tafjord}]{arc-challenge}
Peter Clark, Isaac Cowhey, Oren Etzioni, Tushar Khot, Ashish Sabharwal, Carissa Schoenick, and Oyvind Tafjord. 2018.
\newblock \href {https://arxiv.org/abs/1803.05457} {Think you have solved question answering? try arc, the {AI2} reasoning challenge}.
\newblock \emph{CoRR}, abs/1803.05457.

\bibitem[{Cuconasu et~al.(2024)Cuconasu, Trappolini, Siciliano, Filice, Campagnano, Maarek, Tonellotto, and Silvestri}]{powerofnoise}
Florin Cuconasu, Giovanni Trappolini, Federico Siciliano, Simone Filice, Cesare Campagnano, Yoelle Maarek, Nicola Tonellotto, and Fabrizio Silvestri. 2024.
\newblock \href {https://doi.org/10.1145/3626772.3657834} {The power of noise: Redefining retrieval for rag systems}.
\newblock In \emph{Proceedings of the 47th International ACM SIGIR Conference on Research and Development in Information Retrieval}, SIGIR '24, page 719–729, New York, NY, USA. Association for Computing Machinery.

\bibitem[{Dai et~al.(2022)Dai, Dong, Hao, Sui, Chang, and Wei}]{dai-etal-2022-knowledge}
Damai Dai, Li~Dong, Yaru Hao, Zhifang Sui, Baobao Chang, and Furu Wei. 2022.
\newblock \href {https://doi.org/10.18653/v1/2022.acl-long.581} {Knowledge neurons in pretrained transformers}.
\newblock In \emph{Proceedings of the 60th Annual Meeting of the Association for Computational Linguistics (Volume 1: Long Papers)}, pages 8493--8502, Dublin, Ireland. Association for Computational Linguistics.

\bibitem[{Gao et~al.(2024)Gao, Tow, Abbasi, Biderman, Black, DiPofi, Foster, Golding, Hsu, Le~Noac'h, Li, McDonell, Muennighoff, Ociepa, Phang, Reynolds, Schoelkopf, Skowron, Sutawika, Tang, Thite, Wang, Wang, and Zou}]{eval-harness}
Leo Gao, Jonathan Tow, Baber Abbasi, Stella Biderman, Sid Black, Anthony DiPofi, Charles Foster, Laurence Golding, Jeffrey Hsu, Alain Le~Noac'h, Haonan Li, Kyle McDonell, Niklas Muennighoff, Chris Ociepa, Jason Phang, Laria Reynolds, Hailey Schoelkopf, Aviya Skowron, Lintang Sutawika, and 5 others. 2024.
\newblock \href {https://doi.org/10.5281/zenodo.12608602} {The language model evaluation harness}.

\bibitem[{Geva et~al.(2021)Geva, Schuster, Berant, and Levy}]{geva-etal-2021-transformer}
Mor Geva, Roei Schuster, Jonathan Berant, and Omer Levy. 2021.
\newblock \href {https://doi.org/10.18653/v1/2021.emnlp-main.446} {Transformer feed-forward layers are key-value memories}.
\newblock In \emph{Proceedings of the 2021 Conference on Empirical Methods in Natural Language Processing}, pages 5484--5495, Online and Punta Cana, Dominican Republic. Association for Computational Linguistics.

\bibitem[{Glass et~al.(2022)Glass, Rossiello, Chowdhury, Naik, Cai, and Gliozzo}]{glass-etal-2022-re2g}
Michael Glass, Gaetano Rossiello, Md~Faisal~Mahbub Chowdhury, Ankita Naik, Pengshan Cai, and Alfio Gliozzo. 2022.
\newblock \href {https://doi.org/10.18653/v1/2022.naacl-main.194} {{R}e2{G}: Retrieve, rerank, generate}.
\newblock In \emph{Proceedings of the 2022 Conference of the North American Chapter of the Association for Computational Linguistics: Human Language Technologies}, pages 2701--2715, Seattle, United States. Association for Computational Linguistics.

\bibitem[{Huang et~al.(2025)Huang, Yu, Ma, Zhong, Feng, Wang, Chen, Peng, Feng, Qin, and Liu}]{huang2025survey}
Lei Huang, Weijiang Yu, Weitao Ma, Weihong Zhong, Zhangyin Feng, Haotian Wang, Qianglong Chen, Weihua Peng, Xiaocheng Feng, Bing Qin, and Ting Liu. 2025.
\newblock \href {https://doi.org/10.1145/3703155} {A survey on hallucination in large language models: Principles, taxonomy, challenges, and open questions}.
\newblock \emph{ACM Transactions on Information Systems}, 43(2):1--55.

\bibitem[{Joshi et~al.(2017)Joshi, Choi, Weld, and Zettlemoyer}]{joshi2017triviaqa}
Mandar Joshi, Eunsol Choi, Daniel Weld, and Luke Zettlemoyer. 2017.
\newblock \href {https://doi.org/10.18653/v1/P17-1147} {{T}rivia{QA}: A large scale distantly supervised challenge dataset for reading comprehension}.
\newblock In \emph{Proceedings of the 55th Annual Meeting of the Association for Computational Linguistics (Volume 1: Long Papers)}, pages 1601--1611, Vancouver, Canada. Association for Computational Linguistics.

\bibitem[{Kwiatkowski et~al.(2019)Kwiatkowski, Palomaki, Redfield, Collins, Parikh, Alberti, Epstein, Polosukhin, Devlin, Lee, Toutanova, Jones, Kelcey, Chang, Dai, Uszkoreit, Le, and Petrov}]{kwiatkowski2019natural}
Tom Kwiatkowski, Jennimaria Palomaki, Olivia Redfield, Michael Collins, Ankur Parikh, Chris Alberti, Danielle Epstein, Illia Polosukhin, Jacob Devlin, Kenton Lee, Kristina Toutanova, Llion Jones, Matthew Kelcey, Ming-Wei Chang, Andrew~M. Dai, Jakob Uszkoreit, Quoc Le, and Slav Petrov. 2019.
\newblock \href {https://doi.org/10.1162/tacl_a_00276} {Natural questions: A benchmark for question answering research}.
\newblock \emph{Transactions of the Association for Computational Linguistics}, 7:452--466.

\bibitem[{Lassance et~al.(2024)Lassance, D{\'e}jean, Formal, and Clinchant}]{lassance2024splade}
Carlos Lassance, Herv{\'e} D{\'e}jean, Thibault Formal, and St{\'e}phane Clinchant. 2024.
\newblock Splade-v3: New baselines for splade.
\newblock \emph{arXiv preprint arXiv:2403.06789}.

\bibitem[{Lewis et~al.(2020)Lewis, Perez, Piktus, Petroni, Karpukhin, Goyal, K\"{u}ttler, Lewis, Yih, Rockt\"{a}schel, Riedel, and Kiela}]{lewis2020retrieval}
Patrick Lewis, Ethan Perez, Aleksandra Piktus, Fabio Petroni, Vladimir Karpukhin, Naman Goyal, Heinrich K\"{u}ttler, Mike Lewis, Wen-tau Yih, Tim Rockt\"{a}schel, Sebastian Riedel, and Douwe Kiela. 2020.
\newblock \href {https://proceedings.neurips.cc/paper_files/paper/2020/file/6b493230205f780e1bc26945df7481e5-Paper.pdf} {Retrieval-augmented generation for knowledge-intensive nlp tasks}.
\newblock In \emph{Advances in Neural Information Processing Systems}, volume~33, pages 9459--9474. Curran Associates, Inc.

\bibitem[{Lin et~al.(2022)Lin, Hilton, and Evans}]{lin-etal-2022-truthfulqa}
Stephanie Lin, Jacob Hilton, and Owain Evans. 2022.
\newblock \href {https://doi.org/10.18653/v1/2022.acl-long.229} {{T}ruthful{QA}: Measuring how models mimic human falsehoods}.
\newblock In \emph{Proceedings of the 60th Annual Meeting of the Association for Computational Linguistics (Volume 1: Long Papers)}, pages 3214--3252, Dublin, Ireland. Association for Computational Linguistics.

\bibitem[{Liu and Liu(2023)}]{liu2023memotrap}
Alisa Liu and Jiacheng Liu. 2023.
\newblock \href {https://github.com/inverse-scaling/prize/blob/main/data-release/README.md} {The memotrap dataset}.

\bibitem[{Llama~Team(2024)}]{dubey2024llama3herdmodels}
AI~@~Meta Llama~Team. 2024.
\newblock \href {https://arxiv.org/abs/2407.21783} {The llama 3 herd of models}.
\newblock \emph{Preprint}, arXiv:2407.21783.

\bibitem[{Mallen et~al.(2023)Mallen, Asai, Zhong, Das, Khashabi, and Hajishirzi}]{mallen-etal-2023-trust}
Alex Mallen, Akari Asai, Victor Zhong, Rajarshi Das, Daniel Khashabi, and Hannaneh Hajishirzi. 2023.
\newblock \href {https://doi.org/10.18653/v1/2023.acl-long.546} {When not to trust language models: Investigating effectiveness of parametric and non-parametric memories}.
\newblock In \emph{Proceedings of the 61st Annual Meeting of the Association for Computational Linguistics (Volume 1: Long Papers)}, pages 9802--9822, Toronto, Canada. Association for Computational Linguistics.

\bibitem[{Meng et~al.(2022)Meng, Bau, Andonian, and Belinkov}]{NEURIPS2022_6f1d43d5}
Kevin Meng, David Bau, Alex Andonian, and Yonatan Belinkov. 2022.
\newblock \href {https://proceedings.neurips.cc/paper_files/paper/2022/file/6f1d43d5a82a37e89b0665b33bf3a182-Paper-Conference.pdf} {Locating and editing factual associations in gpt}.
\newblock In \emph{Advances in Neural Information Processing Systems}, volume~35, pages 17359--17372. Curran Associates, Inc.

\bibitem[{Meng et~al.(2023)Meng, Sen~Sharma, Andonian, Belinkov, and Bau}]{meng2022mass}
Kevin Meng, Arnab Sen~Sharma, Alex Andonian, Yonatan Belinkov, and David Bau. 2023.
\newblock Mass editing memory in a transformer.
\newblock \emph{The Eleventh International Conference on Learning Representations (ICLR)}.

\bibitem[{Petroni et~al.(2021)Petroni, Piktus, Fan, Lewis, Yazdani, De~Cao, Thorne, Jernite, Karpukhin, Maillard, Plachouras, Rockt{\"a}schel, and Riedel}]{petroni-etal-2021-kilt}
Fabio Petroni, Aleksandra Piktus, Angela Fan, Patrick Lewis, Majid Yazdani, Nicola De~Cao, James Thorne, Yacine Jernite, Vladimir Karpukhin, Jean Maillard, Vassilis Plachouras, Tim Rockt{\"a}schel, and Sebastian Riedel. 2021.
\newblock \href {https://doi.org/10.18653/v1/2021.naacl-main.200} {{KILT}: a benchmark for knowledge intensive language tasks}.
\newblock In \emph{Proceedings of the 2021 Conference of the North American Chapter of the Association for Computational Linguistics: Human Language Technologies}, pages 2523--2544, Online. Association for Computational Linguistics.

\bibitem[{Rau et~al.(2024)Rau, D{\'e}jean, Chirkova, Formal, Wang, Clinchant, and Nikoulina}]{rau-etal-2024-bergen}
David Rau, Herv{\'e} D{\'e}jean, Nadezhda Chirkova, Thibault Formal, Shuai Wang, St{\'e}phane Clinchant, and Vassilina Nikoulina. 2024.
\newblock \href {https://doi.org/10.18653/v1/2024.findings-emnlp.449} {{BERGEN}: A benchmarking library for retrieval-augmented generation}.
\newblock In \emph{Findings of the Association for Computational Linguistics: EMNLP 2024}, pages 7640--7663, Miami, Florida, USA. Association for Computational Linguistics.

\bibitem[{Shi et~al.(2024{\natexlab{a}})Shi, Jin, Shen, Dong, Wu, and Xiong}]{shi2024ircan}
Dan Shi, Renren Jin, Tianhao Shen, Weilong Dong, Xinwei Wu, and Deyi Xiong. 2024{\natexlab{a}}.
\newblock \href {https://doi.org/10.52202/079017-0162} {Ircan: Mitigating knowledge conflicts in llm generation via identifying and reweighting context-aware neurons}.
\newblock \emph{Advances in Neural Information Processing Systems}, 37:4997--5024.

\bibitem[{Shi et~al.(2023)Shi, Chen, Misra, Scales, Dohan, Chi, Sch{\"a}rli, and Zhou}]{shi2023large}
Freda Shi, Xinyun Chen, Kanishka Misra, Nathan Scales, David Dohan, Ed~H Chi, Nathanael Sch{\"a}rli, and Denny Zhou. 2023.
\newblock \href {https://proceedings.mlr.press/v202/shi23a.html} {Large language models can be easily distracted by irrelevant context}.
\newblock In \emph{Proceedings of the 40th International Conference on Machine Learning}, volume 202, pages 31210--31227. PMLR.

\bibitem[{Shi et~al.(2024{\natexlab{b}})Shi, Han, Lewis, Tsvetkov, Zettlemoyer, and Yih}]{shi-etal-2024-trusting}
Weijia Shi, Xiaochuang Han, Mike Lewis, Yulia Tsvetkov, Luke Zettlemoyer, and Wen-tau Yih. 2024{\natexlab{b}}.
\newblock \href {https://doi.org/10.18653/v1/2024.naacl-short.69} {Trusting your evidence: Hallucinate less with context-aware decoding}.
\newblock In \emph{Proceedings of the 2024 Conference of the North American Chapter of the Association for Computational Linguistics: Human Language Technologies (Volume 2: Short Papers)}, pages 783--791, Mexico City, Mexico. Association for Computational Linguistics.

\bibitem[{Stelmakh et~al.(2022)Stelmakh, Luan, Dhingra, and Chang}]{stelmakh2022asqa}
Ivan Stelmakh, Yi~Luan, Bhuwan Dhingra, and Ming-Wei Chang. 2022.
\newblock \href {https://doi.org/10.18653/v1/2022.emnlp-main.566} {{ASQA}: Factoid questions meet long-form answers}.
\newblock In \emph{Proceedings of the 2022 Conference on Empirical Methods in Natural Language Processing}, pages 8273--8288, Abu Dhabi, United Arab Emirates. Association for Computational Linguistics.

\bibitem[{Sundararajan et~al.(2017)Sundararajan, Taly, and Yan}]{sundararajan2017axiomatic}
Mukund Sundararajan, Ankur Taly, and Qiqi Yan. 2017.
\newblock \href {https://proceedings.mlr.press/v70/sundararajan17a.html} {Axiomatic attribution for deep networks}.
\newblock In \emph{Proceedings of the 34th International Conference on Machine Learning}, volume~70, pages 3319--3328. PMLR.

\bibitem[{Wang and Yu(2025)}]{wang-yu-2025-iquest}
Shuai Wang and Yinan Yu. 2025.
\newblock \href {https://aclanthology.org/2025.acl-long.760/} {i{QUEST}: An iterative question-guided framework for knowledge base question answering}.
\newblock In \emph{Proceedings of the 63rd Annual Meeting of the Association for Computational Linguistics (Volume 1: Long Papers)}, pages 15616--15628, Vienna, Austria. Association for Computational Linguistics.

\bibitem[{Welbl et~al.(2017)Welbl, Liu, and Gardner}]{welbl2017crowdsourcing}
Johannes Welbl, Nelson~F. Liu, and Matt Gardner. 2017.
\newblock \href {https://doi.org/10.18653/v1/W17-4413} {Crowdsourcing multiple choice science questions}.
\newblock In \emph{Proceedings of the 3rd Workshop on Noisy User-generated Text}, pages 94--106, Copenhagen, Denmark. Association for Computational Linguistics.

\bibitem[{Wu et~al.(2026)Wu, Wang, Sun, Lu, Zhang, and Chen}]{conflictrag}
Haiyan Wu, Chenchen Wang, Chaoqun Sun, Chengxiong Lu, Zhiqiang Zhang, and Yanhong Chen. 2026.
\newblock \href {https://doi.org/10.1145/3774904.3792289} {Conflict-aware rag: Multi-stage learning with conflict signals for robust retrieval-augmented generation}.
\newblock In \emph{Proceedings of the ACM Web Conference 2026}, WWW '26, page 2114–2125, New York, NY, USA. Association for Computing Machinery.

\bibitem[{Wu et~al.(2025)Wu, Cai, Yan, Sun, Li, Wang, Yin, and Gao}]{wu-etal-2025-pa}
Jiayi Wu, Hengyi Cai, Lingyong Yan, Hao Sun, Xiang Li, Shuaiqiang Wang, Dawei Yin, and Ming Gao. 2025.
\newblock \href {https://doi.org/10.18653/v1/2025.naacl-long.459} {{PA}-{RAG}: {RAG} alignment via multi-perspective preference optimization}.
\newblock In \emph{Proceedings of the 2025 Conference of the Nations of the Americas Chapter of the Association for Computational Linguistics: Human Language Technologies (Volume 1: Long Papers)}, pages 9091--9112, Albuquerque, New Mexico. Association for Computational Linguistics.

\bibitem[{Xu et~al.(2024)Xu, Shi, and Choi}]{xu2023recomp}
Fangyuan Xu, Weijia Shi, and Eunsol Choi. 2024.
\newblock \href {https://openreview.net/forum?id=mlJLVigNHp} {{RECOMP:} improving retrieval-augmented lms with context compression and selective augmentation}.
\newblock In \emph{The Twelfth International Conference on Learning Representations, {ICLR} 2024, Vienna, Austria, May 7-11, 2024}. OpenReview.net.

\bibitem[{Yang et~al.(2024)Yang, Yang, Zhang, Hui, Zheng, Yu, Li, Liu, Huang, Wei, Lin, Yang, Tu, Zhang, Yang, Yang, Zhou, Lin, Dang, Lu, Bao, Yang, Yu, Li, Xue, Zhang, Zhu, Men, Lin, Li, Tang, Xia, Ren, Ren, Fan, Su, Zhang, Wan, Liu, Cui, Zhang, and Qiu}]{yang2024qwen2}
An~Yang, Baosong Yang, Beichen Zhang, Binyuan Hui, Bo~Zheng, Bowen Yu, Chengyuan Li, Dayiheng Liu, Fei Huang, Haoran Wei, Huan Lin, Jian Yang, Jianhong Tu, Jianwei Zhang, Jianxin Yang, Jiaxi Yang, Jingren Zhou, Junyang Lin, Kai Dang, and 23 others. 2024.
\newblock Qwen2.5 technical report.
\newblock \emph{arXiv preprint arXiv:2412.15115}.

\bibitem[{Yang et~al.(2018)Yang, Qi, Zhang, Bengio, Cohen, Salakhutdinov, and Manning}]{yang2018hotpotqa}
Zhilin Yang, Peng Qi, Saizheng Zhang, Yoshua Bengio, William Cohen, Ruslan Salakhutdinov, and Christopher~D. Manning. 2018.
\newblock \href {https://doi.org/10.18653/v1/D18-1259} {{H}otpot{QA}: A dataset for diverse, explainable multi-hop question answering}.
\newblock In \emph{Proceedings of the 2018 Conference on Empirical Methods in Natural Language Processing}, pages 2369--2380, Brussels, Belgium. Association for Computational Linguistics.

\bibitem[{Yoran et~al.(2024)Yoran, Wolfson, Ram, and Berant}]{yoran2024makingretrievalaugmentedlanguagemodels}
Ori Yoran, Tomer Wolfson, Ori Ram, and Jonathan Berant. 2024.
\newblock \href {https://openreview.net/forum?id=ZS4m74kZpH} {Making retrieval-augmented language models robust to irrelevant context}.
\newblock In \emph{The Twelfth International Conference on Learning Representations, {ICLR} 2024, Vienna, Austria, May 7-11, 2024}. OpenReview.net.

\bibitem[{Zellers et~al.(2019)Zellers, Holtzman, Bisk, Farhadi, and Choi}]{zellers-etal-2019-hellaswag}
Rowan Zellers, Ari Holtzman, Yonatan Bisk, Ali Farhadi, and Yejin Choi. 2019.
\newblock \href {https://doi.org/10.18653/v1/P19-1472} {{H}ella{S}wag: Can a machine really finish your sentence?}
\newblock In \emph{Proceedings of the 57th Annual Meeting of the Association for Computational Linguistics}, pages 4791--4800, Florence, Italy. Association for Computational Linguistics.

\bibitem[{Zhang et~al.(2026{\natexlab{a}})Zhang, Zhang, Pang, Tong, Zheng, and Zheng}]{zhang2026less}
Qianchi Zhang, Hainan Zhang, Liang Pang, Yongxin Tong, Hongwei Zheng, and Zhiming Zheng. 2026{\natexlab{a}}.
\newblock \href {https://doi.org/10.1145/3774904.3792158} {Less is more: Compact clue selection for efficient retrieval-augmented generation reasoning}.
\newblock In \emph{Proceedings of the ACM Web Conference 2026}, WWW '26, page 1971–1982, New York, NY, USA. Association for Computing Machinery.

\bibitem[{Zhang et~al.(2026{\natexlab{b}})Zhang, Li, Huang, Cheng, Guo, Zhu, Wang, Wang, and Xiao}]{zhang2026sentineldecodingcontextutilization}
Yong Zhang, Heng Li, Yanwen Huang, Ning Cheng, Yang Guo, Yun Zhu, Yanmeng Wang, Shaojun Wang, and Jing Xiao. 2026{\natexlab{b}}.
\newblock \href {https://arxiv.org/abs/2505.23277} {Sentinel: Decoding context utilization via attention probing for efficient llm context compression}.
\newblock \emph{Preprint}, arXiv:2505.23277.

\bibitem[{Zhao et~al.(2026)Zhao, Huang, Zhong, Hu, Zhang, Hu, and Zhang}]{zhao-etal-2026-learning-extract}
Xinping Zhao, Shouzheng Huang, Yan Zhong, Xinshuo Hu, Meishan Zhang, Baotian Hu, and Min Zhang. 2026.
\newblock \href {https://doi.org/10.18653/v1/2026.findings-acl.782} {Learning to extract rational evidence via reinforcement learning for retrieval-augmented generation}.
\newblock In \emph{Findings of the {A}ssociation for {C}omputational {L}inguistics: {ACL} 2026}, pages 15934--15956, San Diego, California, United States. Association for Computational Linguistics.

\bibitem[{Zhou et~al.(2023)Zhou, Zhang, Poon, and Chen}]{zhou-etal-2023-context}
Wenxuan Zhou, Sheng Zhang, Hoifung Poon, and Muhao Chen. 2023.
\newblock \href {https://doi.org/10.18653/v1/2023.findings-emnlp.968} {Context-faithful prompting for large language models}.
\newblock In \emph{Findings of the Association for Computational Linguistics: EMNLP 2023}, pages 14544--14556, Singapore. Association for Computational Linguistics.

\end{thebibliography}

\appendix

\newpage
\section{Appendix}

\subsection{Neuron Mining Dataset Setting}
\label{app:neuron selection}
For neuron mining, we use the first 100 samples from the HotpotQA training distractor split~\cite{yang2018hotpotqa}. To simulate a realistic RAG setting where multiple retrieved documents are provided as input, we convert each sample into the context-provided prompt format shown in Table~\ref{tb:input_format}, where all documents in the sample are directly used as the input context without an additional retriever. The corresponding prompt format without retrieved context is shown in Table~\ref{tb:attr_format}.

\begin{table}[h]
\centering
\begin{tcolorbox}[
    title={\small\textbf{Input Format With Retrieved Context}},
    colback=gray!5,
    colframe=gray!60,
    fonttitle=\bfseries,
    boxrule=0.5pt,
    arc=3pt
]
{\small
\textbf{System:} Your task is to extract relevant information from provided documents and to answer questions as briefly as possible.\\
\textbf{User:}\\
Background:\\
Document 1: [title] [sentences]\\
Document 2: [title] [sentences]\\
\hspace*{1em}$\vdots$\\
Document N: [title] [sentences]\\[0.4em]
Question: [question]
}
\end{tcolorbox}
\caption{Input prompt format with retrieved context, used for attribution score computation.}
\label{tb:input_format}
\end{table}

\begin{table}[h]
\centering
\begin{tcolorbox}[
    title={\small\textbf{Input Format Without Retrieved Context}},
    colback=gray!5,
    colframe=gray!60,
    fonttitle=\bfseries,
    boxrule=0.5pt,
    arc=3pt
]
{\small
{
\textbf{System:} Answer the questions as briefly as possible.\\[0.3em]
\textbf{User:}
Question: [question]
}
}
\end{tcolorbox}
\caption{Input prompt format without retrieved context, used for attribution score computation.}
\label{tb:attr_format}
\end{table}

\subsection{Baseline Implementation Details}
To ensure a consistent comparison, for \code{PA-RAG}, we use the authors’ publicly released Llama-3-8B-Instruct checkpoint and reproduce the method on Qwen-2.5-7B-Instruct using their training recipe.  
For \code{RetRobust}, we reproduce results on LLaMA-3-8B-Instruct and Qwen-2.5-7B-Instruct using the authors' training recipe. 
For \code{CAD}, we set $\alpha=0.5$. 
For \code{IRCAN}, we use the same backbone as \alg and select the top 20 neurons by attribution as the candidate pool, with $\alpha=7$ and $k=5$ final neurons.

\subsection{Inference-Time RAG Prompt}
The prompt format used for inference-time RAG generation is shown in Table~\ref{tb:inference_format}. The ``Background'' section consists of the top-5 documents retrieved by the retriever.

\begin{table}[t]
\centering
\begin{tcolorbox}[
    title={\small\textbf{Inference Input Format}},
    colback=gray!5,
    colframe=gray!60,
    fonttitle=\bfseries,
    boxrule=0.5pt,
    arc=3pt
]
{\small
\textbf{System:} You are a helpful assistant. Your task is to extract relevant information from provided documents and to answer questions as briefly as possible.\\[0.3em]
\textbf{User:}\\
Background:\\
Document 1: ...\\
Document 2: ...\\
\hspace*{1em}$\vdots$\\
Document 5: ...\\[0.4em]
Question: [question]
}
\end{tcolorbox}
\caption{Input prompt format used for inference-time RAG.}
\label{tb:inference_format}
\end{table}

\subsection{LLM-based Evaluation}
\label{app:llm_eval}
We evaluate model outputs using GPT-5-mini as an LLM judge on NQ, SCIQ, and HotpotQA with Llama-3-8B-Instruct. We use the LLM-as-a-judge evaluation protocol provided by \citet{rau-etal-2024-bergen}. 
As shown in Table~\ref{tab:llm_eval}, \alg achieves the highest average score and the best performance on two of the three datasets. 

\begin{table}[t]
\centering
\small
\begin{tabular}{lccc}
\toprule
\textbf{Method} & \textbf{SCIQ} & \textbf{NQ} & \textbf{HQA} \\
\midrule
\code{RAG}          & 67.00 & 57.03 & 52.14 \\
\code{PA-RAG}       & 60.80 & 55.66 & 52.21 \\
\code{CAD}          & 59.40 & 56.75 & 47.12 \\
\code{IRCAN}        & 67.80 & 57.81 & \textbf{54.70} \\
\alg (Ours) & \textbf{67.90} & \textbf{58.69} & 54.39 \\
\bottomrule
\end{tabular}
\caption{LLM-based evaluation results on Llama-3-8B-Instruct using GPT-5-mini as the judge.}
\label{tab:llm_eval}
\end{table}

\subsection{Random Neuron Selection}
\label{app:random_neuron}
To test whether \alg’s gains depend on neuron selection, we select five random neurons and apply the identical enhancement strength on Llama-3-8B-Instruct. 
As shown in Table~\ref{tab:random_neuron}, random neuron selection yields performance comparable to vanilla RAG across all three datasets.
In contrast, \alg, using the same strength, achieves the best performance across datasets, suggesting that neuron selection drives the gain.

\begin{table}[t]
\centering
\small
\begin{tabular}{lcccc}
\toprule
\textbf{Method} & \textbf{NQ} & \textbf{SCIQ} & \textbf{HQA} & \textbf{Avg.} \\
\midrule
Random (seed 42)  & 62.71 & 54.10 & 46.68 & 54.50 \\
Random (seed 77)  & 62.53 & 53.90 & 46.70 & 54.38 \\
Random (seed 99)  & 62.46 & 54.10 & 46.48 & 54.35 \\
Random (seed 512) & 62.99 & 53.80 & 46.45 & 54.41 \\
Random (seed 256) & 62.81 & 53.90 & 46.54 & 54.42 \\
\midrule
RAG & 62.81 & 54.10 & 46.55 & 54.49 \\
\alg & \textbf{66.44} & \textbf{57.10} & \textbf{52.27} & \textbf{58.60} \\
\bottomrule
\end{tabular}
\caption{Random-neuron control experiment on Llama-3-8B-Instruct. Each random selection uses the same number of neurons $k=5$ and enhancement strength $\alpha=7$ as \alg.}
\label{tab:random_neuron}
\end{table}

\subsection{Sensitivity to Mining Samples}
\label{app:mining_samples}
We show that \alg maintains its performance across different sets of 100 examples used for mining. 
The seed only determines which 100 examples are drawn from the HotpotQA training split. 
We therefore mined neurons with various random seeds on Llama-3-8B-Instruct and evaluated on HotpotQA. As shown in Table~\ref{tab:mining_samples}, match accuracy remains stable across seeds ($52.69 \pm 0.44$).
\begin{table}[t]
\centering
\small
\begin{tabular}{lc}
\toprule
\textbf{Mining Set} & \textbf{Accuracy} \\
\midrule
First 100 (Ours) & 52.27 \\
Seed 11  & 53.75 \\
Seed 55  & 53.14 \\
Seed 741 & 52.77 \\
Seed 333 & 52.64 \\
Seed 512 & 52.54 \\
Seed 150 & 52.50 \\
Seed 421 & 52.48 \\
Seed 107 & 52.46 \\
Seed 909 & 52.39 \\
\midrule
Avg. $\pm$ Std. & $52.69 \pm 0.44$ \\
\bottomrule
\end{tabular}
\caption{Sensitivity to the choice of neuron-mining examples on HotpotQA using Llama-3-8B-Instruct. The first row uses the initial 100 training examples, while the remaining rows use 100 examples randomly sampled with different seeds.}
\label{tab:mining_samples}
\end{table}

\subsection{Effect of Enhancement Strength and Number of Selected Neurons}
\label{effect_neuron}

\begin{table}[t]
\centering
\resizebox{1.00\columnwidth}{!}{
\begin{tabular}{ccccccc}
\thickhline
& \multicolumn{2}{c}{\textbf{NQ}}
& \multicolumn{2}{c}{\textbf{SCIQ}}
& \multicolumn{2}{c}{\textbf{HQA}} \\
\cmidrule(lr){2-3} \cmidrule(lr){4-5} \cmidrule(lr){6-7}
\textbf{$\alpha$} & $k{=}5$ & $k{=}15$ & $k{=}5$ & $k{=}15$ & $k{=}5$ & $k{=}15$ \\
\hline
3 & 63.80 & 63.27 & 53.50 & 54.00 & 49.91 & 49.57 \\
5 & 65.81 & 65.67 & 56.20 & 58.00 & 52.20 & 52.66 \\
7 & 66.44 & 66.02 & 57.10 & 57.70 & 52.27 & 52.09 \\
\thickhline
\end{tabular}
}
\vspace{-6pt}
\caption{
  Effect of enhancement strength $\alpha$ and top-$k$ neuron mining.
}
\vspace{-8pt}
\label{tab:alpha_topk}
\end{table}

We additionally evaluate different enhancement strengths $\alpha \in \{3,5,7\}$ and top-$k$ values $k \in \{5,15\}$ on NQ, HotpotQA, and SCIQ using Llama-3-8B-Instruct. 
Larger enhancement strengths generally lead to better performance, with $\alpha=7$ achieving the strongest overall results. We also observe that $k=5$ and $k=15$ yield comparable performance, suggesting that the neurons most important for selective context utilization are already concentrated within a small set of top-ranked neurons.

\subsection{Effect of Candidate Pool Size}
\label{pool size}
To validate the choice of 50 candidate neurons, we vary the size of $A^{(t)}$ and $B^{(t)}$ over $\{20, 50, 80\}$ before taking their intersection.
Table~\ref{tab:candidate_pool} reports the results on HotpotQA, SCIQ, and NQ using Llama-3-8B-Instruct and Qwen-2.5-7B-Instruct.

\begin{table}[t]
\centering
\small
\begin{tabular}{lccc}
\toprule
\textbf{Pool Size} & \textbf{HQA} & \textbf{SCIQ} & \textbf{NQ} \\
\midrule
\multicolumn{4}{c}{\textit{Llama-3-8B-Instruct}} \\
\cmidrule(lr){1-4}
Top-20 & \textbf{52.46} & \textbf{57.90} & \textbf{66.55} \\
Top-50 & 52.27 & 57.10 & 66.44 \\
Top-80 & 50.25 & 54.70 & 63.52 \\

\midrule

\multicolumn{4}{c}{\textit{Qwen-2.5-7B-Instruct}} \\
\cmidrule(lr){1-4}
Top-20 & 46.04 & 53.90 & 62.50 \\
Top-50 & \textbf{47.00} & \textbf{56.20} & \textbf{63.24} \\
Top-80 & 46.04 & 53.90 & 62.50 \\

\bottomrule
\end{tabular}
\caption{Effect of candidate pool size. }
\label{tab:candidate_pool}
\end{table}

We observe that Top-20 is marginally higher than Top-50 by 0.1-0.8 points on Llama-3-8B-Instruct, but on Qwen-2.5-7B-Instruct, Top-50 is best on all three datasets by 0.7-2.3 points, so Top-50 offers the best overall trade-off. Top-20 and top-80 select the same final neuron set; therefore, they yield identical scores.

\subsection{Effect of Context Window Size}
\label{effect window}

\begin{figure}[t]
    \centering
    \includegraphics[width=\columnwidth]{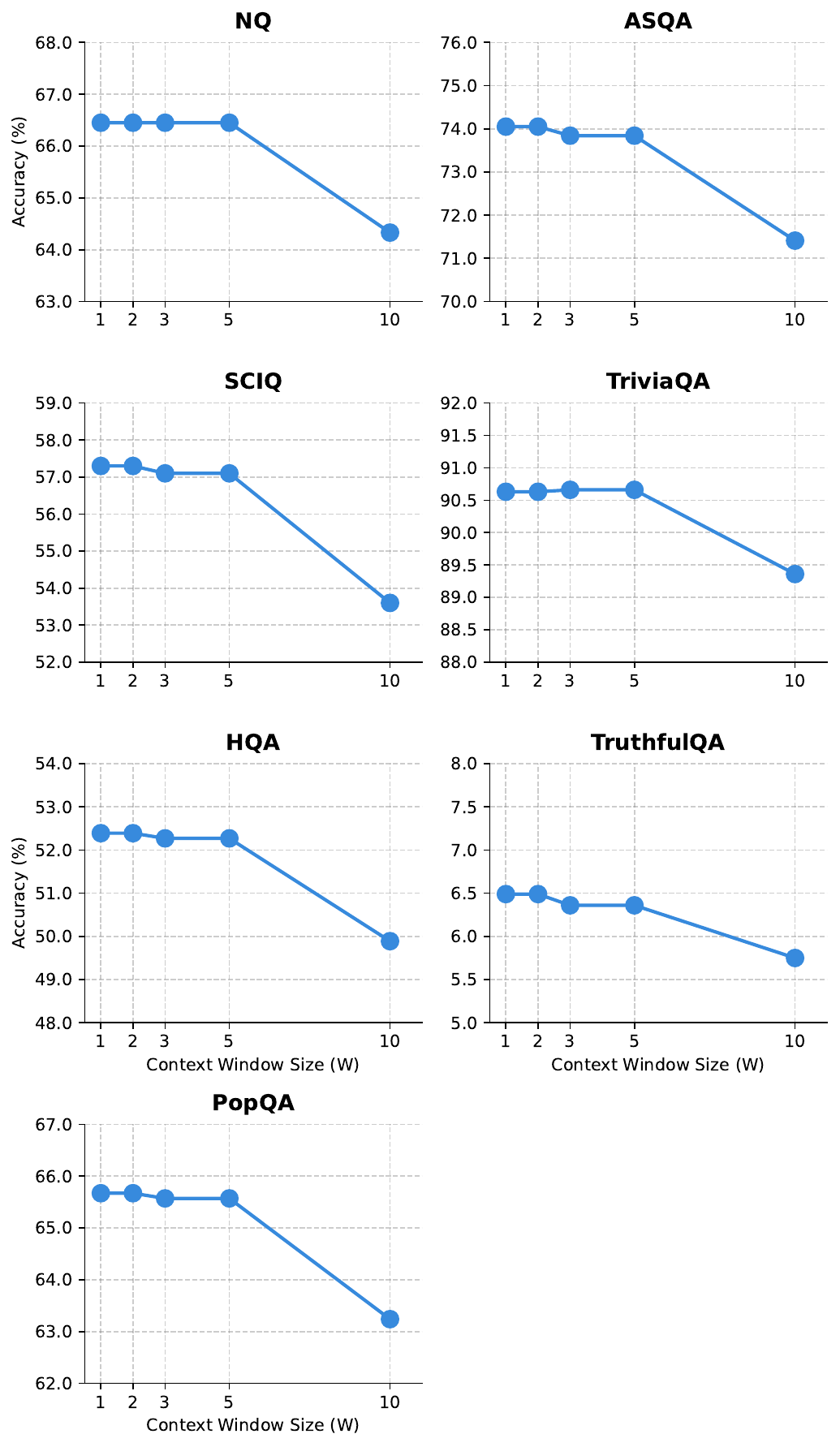}
    \vspace{-2mm}
    \caption{
    Performance sensitivity to context window size $W$.
    }
    \label{fig:window_size}
    \vspace{-2mm}
\end{figure}

We vary the context window size $W \in \{1,2,3,5,10\}$ used to compute attribution variability. As shown in Figure~\ref{fig:window_size}, performance remains stable across small and moderate window sizes ($W=1,2,3,5$), while a large window ($W=10$) consistently degrades performance across datasets.
We additionally observe that the mined neuron sets are highly similar across different window sizes. In particular, the selected neurons for $W=1$ and $W=2$ are identical, and those for $W=3$ and $W=5$ are also identical. Even for $W=10$, the selected neuron set differs from the other settings by at most one neuron. 
Despite these highly similar neuron sets, performance differences remain relatively small for practical window sizes, with degradation mainly observed at $W=10$.

\begin{table}[t]
\centering
\resizebox{0.4\textwidth}{!}{
\begin{tabular}{cccccc}
\thickhline
\textbf{\# Samples} & \textbf{NQ} & \textbf{SCIQ} & \textbf{HQA} & \textbf{Overlap} \\
\hline
100  & 66.44 & 57.10 & 52.27 & -- \\
500  & 66.94 & 57.50 & 52.63 & 3/5 \\
1000 & 64.61 & 54.10 & 50.28 & 3/5 \\
\thickhline
\end{tabular}}
\vspace{-6pt}
\caption{
Effect of neuron mining dataset size on HotpotQA using LLaMA-3-8B-Instruct. Overlap indicates the percentage of shared neurons compared to the 100-sample setting.
}
\vspace{-8pt}
\label{tab:sample_size}
\end{table}

\subsection{Effect of Neuron Mining Dataset Size}
\label{effect dataset}
We additionally analyze the effect of neuron-mining dataset size by selecting neurons using $N \in \{100,500,1000\}$ samples from HotpotQA on Llama-3-8B-Instruct. Table~\ref{tab:sample_size} reports the performance on NQ, SCIQ, and HotpotQA.

We observe that increasing the number of neuron-mining samples does not necessarily improve downstream performance. In particular, $N=100$ and $N=500$ yield comparable results, while performance drops when using $N=1000$, despite requiring substantially more mining data. 
The selected neurons also exhibit considerable overlap across different sample sizes. These results suggest that context-aware neurons can be identified with relatively small mining sets.

\begin{table}[t]
\centering
\small
\begin{tabular}{llcccc}
\toprule
\textbf{Gold Pos.} & \textbf{Method} &
\multicolumn{4}{c}{\textbf{\# Distractors ($N$)}} \\
\cmidrule(lr){3-6}
& & \textbf{0} & \textbf{2} & \textbf{4} & \textbf{8} \\
\midrule

\multirow{3}{*}{First}
& \code{RAG}   & 81.0 & 79.4 & 77.8 & 77.4 \\
& \alg & 84.2 & 83.6 & 82.2 & 81.9 \\
& $\Delta$ & +3.2 & +4.2 & +4.4 & \textbf{+4.5} \\
\midrule

\multirow{3}{*}{Shuffle}
& \code{RAG}   & 81.0 & 79.2 & 77.4 & 75.2 \\
& \alg  & 84.2 & 83.6 & 81.4 & 80.5 \\
& $\Delta$ & +3.2 & +4.4 & +4.0 & \textbf{+5.3} \\
\midrule

\multirow{3}{*}{Last}
& \code{RAG}   & 81.0 & 79.8 & 78.0 & 75.7 \\
& \alg  & 84.2 & 82.5 & 83.1 & 81.6 \\
& $\Delta$ & +3.2 & +2.7 & +5.1 & \textbf{+5.9} \\
\bottomrule
\end{tabular}
\caption{Controlled noise experiment on 1000 HotpotQA validation examples across different gold-document positions. Each context contains two gold documents and $N$ distractors. }
\label{tab:controlled_noise}
\end{table}

\subsection{Validation of Selective Context-Aware Neurons.}
\label{app:neuron_validation}
We hypothesize that a selective context-aware neuron should respond systematically to the composition of retrieved evidence. As gold documents are progressively replaced by distractors, its activation should also change progressively. Accordingly, the activation under the mixed condition should lie between those under the gold-only and distractor-only conditions.
To validate this hypothesis, we analyze the selected neurons on $n$ held-out HotpotQA samples. We use two sample sizes, $n=1,000$ and $5,000$. 
For each neuron, we measure the mean activation at the final input position before answer generation under three context settings: GG, containing two gold documents; GD, containing one gold and one distractor; and DD, containing two distractors.

Table~\ref{tab:neuron_activation} shows that across both evaluation sizes, four of the five neurons exhibit a graded activation pattern in which GD lies between GG and DD.
This indicates that their activations systematically track the composition of supporting and distracting evidence rather than responding uniformly to retrieved context.
The activation patterns and $|\mathrm{GG}-\mathrm{DD}|$ gaps remain nearly unchanged between $n=1,000$ and $n=5,000$, indicating that the observed patterns are stable and are not driven by a small evaluation sample.
These results suggest that cross-input variability-based mining identifies neurons that selectively respond to the composition of retrieved evidence, supporting their characterization as selective context-aware neurons.

\subsection{Controlled Noise Experiments}
\label{app:controlled_noise}
We evaluate robustness under controlled levels of noise. 
On the HotpotQA validation split, we construct each context with 2 gold documents and $N$ distractor documents, where $N \in \{0,2,4,8\}$. 
We keep the same selected neurons and the same 1000 samples across all noise levels, and we vary the gold documents’ position: first, shuffle, last.
Table~\ref{tab:controlled_noise} shows that vanilla RAG drops steadily as distractors are added.
\alg degrades more slowly, and its gain grows with noise compared to vanilla \code{RAG}. 
This indicates that \alg mitigates performance degradation under accumulating distractors. The effect remains largely invariant to the position of gold documents, suggesting that the improvement is not sensitive to gold-document position.

\begin{table}[t]
\centering
\small
\setlength{\tabcolsep}{4pt}
\begin{tabular}{lcccc}
\toprule
\textbf{Neuron} & \textbf{GG} & \textbf{GD} & \textbf{DD} & $\boldsymbol{|\mathrm{GG}-\mathrm{DD}|}$ \\
\midrule
\multicolumn{5}{l}{\textit{$n=1{,}000$}} \\
30@3382  &  6.848 &  6.387 &  5.887 & 0.961 \\
27@8140  & -3.373 & -3.433 & -3.513 & 0.140 \\
30@5035  &  0.188 &  0.168 &  0.163 & 0.025 \\
21@12666 & -0.152 & -0.122 & -0.124 & 0.028 \\
13@2158  & -1.568 & -1.415 & -1.170 & 0.398 \\
\midrule
\multicolumn{5}{l}{\textit{$n=5{,}000$}} \\
30@3382  &  6.870 &  6.392 &  5.900 & 0.970 \\
27@8140  & -3.380 & -3.430 & -3.495 & 0.115 \\
30@5035  &  0.187 &  0.168 &  0.164 & 0.024 \\
21@12666 & -0.153 & -0.123 & -0.124 & 0.029 \\
13@2158  & -1.563 & -1.412 & -1.176 & 0.387 \\
\bottomrule
\end{tabular}
\caption{Mean activations of the five selected neurons under gold-only (GG), mixed gold-distractor (GD), and distractor-only (DD) contexts. The notation $l@i$ denotes the $i$-th intermediate neuron in the $l$-th FFN layer.}
\label{tab:neuron_activation}
\end{table}

\subsection{Evaluation on Out-of-Domain Tasks.}
We evaluate \alg on three out-of-domain tasks using the lm-evaluation-harness to assess whether neuron enhancement affects performance beyond QA: HellaSwag~\cite{zellers-etal-2019-hellaswag}, ARC-Challenge~\cite{arc-challenge}, and MemoTrap~\cite{liu2023memotrap}. Table~\ref{tab:ood} compares \alg with vanilla \code{RAG} and \code{IRCAN}.
On HellaSwag and ARC-Challenge, both \code{IRCAN} and \alg remain close to \code{RAG}, with only small changes on both backbones. 
On MemoTrap, \alg improves clearly over both \code{RAG} and \code{IRCAN} on Llama-3-8B-Instruct. 
On Qwen-2.5-7B-Instruct, where \code{RAG} already performs strongly, all edited variants remain close to it. 
Overall, neuron enhancement does not cause severe degradation of general ability, and its effect varies by task rather than uniformly harming out-of-domain performance. 

\subsection{Experimental Details of Out-of-Domain Tasks}
Out-of-Domain results are obtained with the Eluther AI LM Evaluation Harness~\cite{eval-harness}. 
HellaSwag and ARC-Challenge are run 5-shot and scored by length-normalized accuracy (\texttt{acc\_norm}), while MemoTrap is run zero-shot and scored by accuracy (\texttt{acc}).

\begin{table}[t]
\centering
\resizebox{\columnwidth}{!}{
\begin{tabular}{lccc}
\toprule
\textbf{Method} & \textbf{HellaSwag} & \textbf{ARC-Challenge} & \textbf{MemoTrap} \\
\midrule
\multicolumn{4}{l}{\textit{Llama-3-8B-Instruct}} \\ \hline
\code{RAG} & 78.19 & 62.20 & 49.15 \\
\code{IRCAN}
& 77.04 {(-1.15)}
& 60.58 {(-1.62)}
& 58.12 {(+8.97)} \\
\alg (Ours)
& 76.55 {(-1.64)}
& 59.04 {(-3.16)}
& 65.71 { (+16.56)} \\
\midrule
\multicolumn{4}{l}{\textit{Qwen-2.5-7B-Instruct}} \\ \hline
\code{RAG} & 81.00 & 65.70 & 66.56 \\
\code{IRCAN}
& 81.00 {(0.00)}
& 66.13 {(+0.43)}
& 66.99 {(+0.43)} \\
\alg (Ours)
& 80.61 {(-0.39)}
& 65.27 {(-0.43)}
& 68.06 {(+1.50)} \\
\bottomrule
\end{tabular}
}
\caption{Accuracy comparison across out-of-domain benchmarks using Llama-3-8B-Instruct (top) and Qwen-2.5-7B-Instruct (bottom).}
\label{tab:ood}
\end{table}

\end{document}